\documentclass[letterpaper, 10 pt, conference]{ieeeconf}  % Comment this line out if you need a4paper

\IEEEoverridecommandlockouts                              % This command is only needed if 
\usepackage{graphics} % for pdf, bitmapped graphics files
\usepackage{epsfig} % for postscript graphics files
\usepackage{mathptmx} % assumes new font selection scheme installed
\usepackage{times} % assumes new font selection scheme installed
\usepackage{amsmath} % assumes amsmath package installed
\usepackage{amssymb}  % assumes amsmath package installed
\usepackage{booktabs}
\usepackage[font=footnotesize]{caption}
\usepackage{titlesec}
\usepackage{subcaption}
\title{\textbf{FlockDiffusion:
Assignment-Conditioned Diffusion for Multi-Drone Task Allocation and Completion}}

\author{Iana Zhura$^{*}$, Satenik Akopyan$^{*}$, Roohan Ahmed Khan, \\ Miguel Altamirano Cabrera, Aleksey~Fedoseev and Dzmitry Tsetserukou%
 \thanks{$^{*}$Equal contribution.}%
 \thanks{All authors are with the Skolkovo Institute of Science and Technology, Moscow, Russia. Email: {\tt\small \{ yana.zhura, satenik.akopyan, roohan.khan, m.altamirano, aleksey.fedoseev, d.tsetserukou\}@skoltech.ru}}%
}

\begin{document}

\maketitle
\thispagestyle{empty}
\pagestyle{empty}

%%%%%%%%%%%%%%%%%%%%%%%%%%%%%%%%%%%%%%%%%%%%%%%%%%%%%%%%%%%%%%%%%%%%%%%%%%%%%%%%
\begin{abstract}
Autonomous multi-drone navigation requires fleets to service distributed objectives in cluttered environments under tight computational budgets. Efficient coordination depends on task bundling, where each drone visits multiple objectives along its route. Separate solvers for cost estimation, assignment, and execution incur redundant graph search and produce long, abrupt paths. We propose FlockDiffusion, a learned framework combining a scene graph encoder, an explicit allocation head, an assignment conditioned diffusion transformer, and a closed form trajectory decoder. An autoregressive teacher provides offline supervision for parallel fleet trajectory generation. PyBullet ablations show that bundling increases task completion from 50\% to 100\%, while our complete teacher further reduces route cost by 8.4\% relative to MAGNNET with bundling. In the optimized scalability benchmark, evaluated on 100 scenes per density with ten drones, FlockDiffusion achieves 6.2 to 7.6 times faster inference and approximately 37\% shorter routes than the classical pipeline. As nominal task counts increase from 20 to 40, latency rises from 7.8 to 11.1\,ms, compared with 48.0 to 75.8\,ms for the baseline. In a separate evaluation across five Gazebo environments, FlockDiffusion achieves 100\% planner coverage and reduces planned route cost by 15.4\% relative to the baseline with bundling. These results demonstrate efficient planning under increasing task density in configurations that are demanding to reproduce with physical drone fleets.

\end{abstract}

% ----------------------------------------------------------------------
\section{INTRODUCTION}
\label{sec:intro}
 \begin{figure}[t]
    \centering
    \includegraphics[width=\columnwidth]{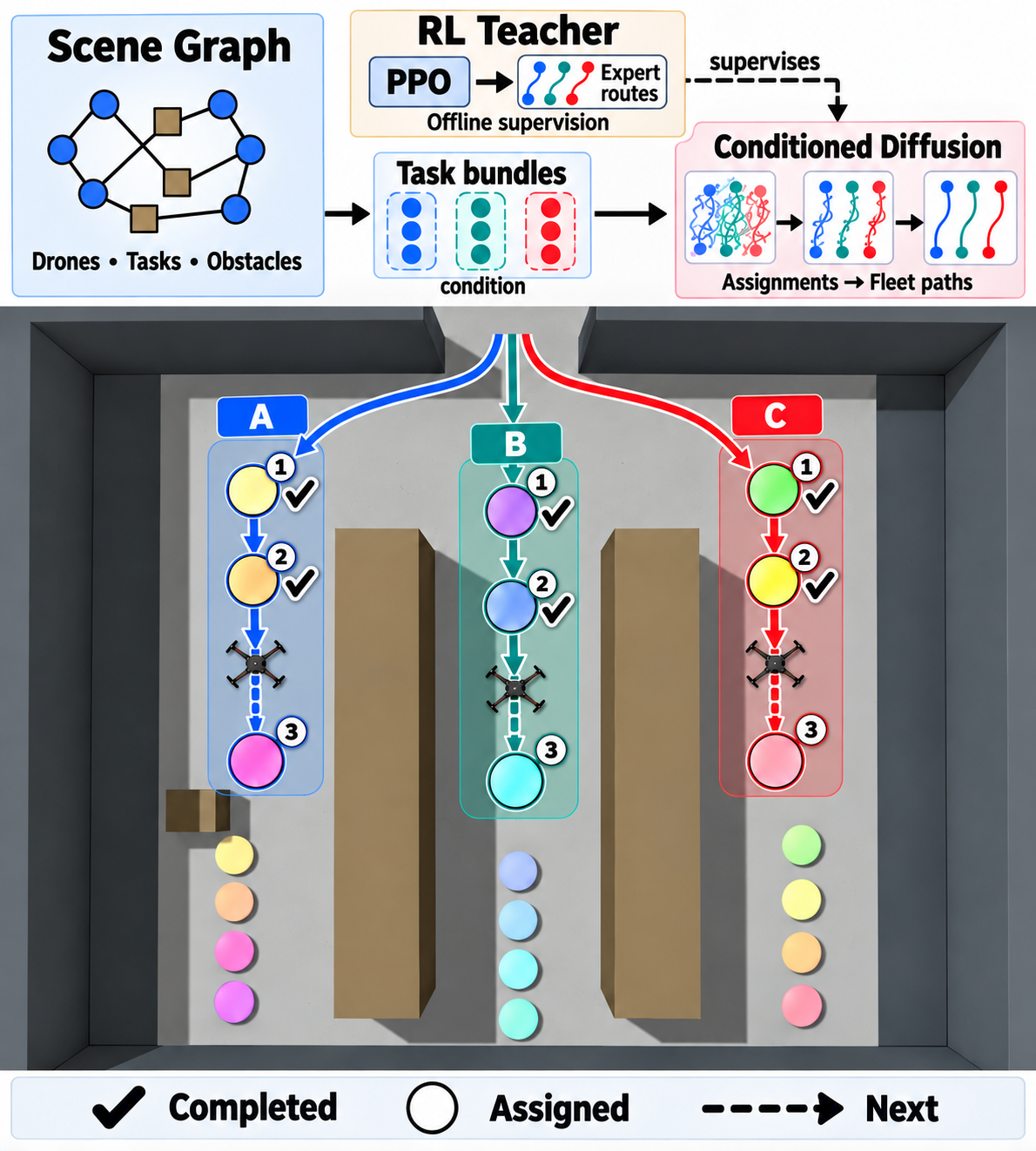}
    \caption{\textbf{FlockDiffusion overview.}
A scene GNN predicts task assignments that are refined into ordered bundles and conditions the parallel diffusion of smooth fleet trajectories. An autoregressive CTDE PPO teacher and route rollout provide offline supervision. Colors associate each drone with tasks collected along its route; check marks denote completed tasks, and dashed curves show the remaining paths.}
    \label{fig:flockdiffusion_teaser}
\end{figure}
The deployment of autonomous drone swarms has grown rapidly across applications such as aerial inspection, disaster response, environmental monitoring, and last-mile delivery~\cite{saunders2024autonomous}. A central bottleneck is assigning tasks to the most suitable agents, especially when tasks outnumber the fleet and each drone must serve several objectives. Effective coordination then depends on \emph{bundling}: a drone should collect multiple tasks that lie along its route rather than completing one before accepting the next. Bundling couples assignment to flight geometry, since whether a task is along the way depends on the obstacle-aware path a drone would actually fly.
 
% Recent learning-based methods combine multi-agent reinforcement learning with graph neural networks to make decentralized allocation decisions from local observations and agent-to-task relations~\cite{ratnabala2025hippo}. The prior system follows this line, pairing a graph neural network encoder with a Proximal Policy Optimization policy and relying on classical planners for cost geometry~\cite{zhang2026cooperative}. Yet such systems share a plan and then fly structure at deployment: an obstacle-aware cost oracle estimates travel costs, a learned policy bids and assigns, and a per drone A star planner computes the executed paths. This has three drawbacks: latency grows with task density, as the oracle graph and A star segments scale with the task count; obstacles are reasoned about twice, once to decide bundling and again to realize it; and executed paths are grid-aligned staircases that are long and abrupt, yielding high jerk trajectories unnatural to fly. These are structural, following from separating assignment from the generation of flight geometry.

Recent learning-based methods combine multi-agent reinforcement learning with graph neural networks to make decentralized allocation decisions from local observations and agent-to-task relations~\cite{ratnabala2025hippo}. MAGNNET~\cite{ratnabala2025magnnet}, which forms the basis of our offline allocation teacher, pairs a graph neural network with a Proximal Policy Optimization policy while relying on separately computed motion costs and classical planning for trajectory execution. More broadly, task allocation and motion planning have also been coupled through reinforcement learning and optimization, but they commonly retain distinct decision and trajectory-generation stages. This separation can introduce repeated geometric computation: obstacle-aware costs are evaluated for allocation, after which a trajectory planner again reasons about the environment to realize the assigned routes. FlockDiffusion instead uses the predicted task allocation and visit order as an explicit conditioning signal for learned fleet trajectory generation.
 
% Denoising diffusion models are powerful generators of continuous outputs~\cite{ho2020ddpm,song2021ddim} and have successfully generated trajectories conditioned on visual observations or LiDAR point clouds~\cite{liang2024dtg,zhura2026embodieddiffusion}. Scene graphs, however, remain comparatively unexplored as conditions for diffusion trajectory generation, despite proving effective for multi-robot task allocation ~\cite{ratnabala2025magnnet,goarin2024gnn}. FlockDiffusion combines these strengths: a graph neural network encodes drones, tasks, and obstacles, predicts task bundles, and conditions a diffusion transformer that generates all drone trajectories in parallel. Few-step DDIM sampling, parallel fleet denoising, and the absence of online graph search enable fast inference.

Denoising diffusion models are powerful generators of continuous outputs~\cite{ho2020ddpm,song2021ddim} and have been applied to robot trajectory generation from visual observations and LiDAR point clouds~\cite{liang2024dtg,zhura2026embodieddiffusion}. Recent work has also extended diffusion to multi-robot motion planning and swarm trajectory generation~\cite{shaoul2025multirobot,ding2025swarmdiff}. These approaches demonstrate that diffusion can generate coordinated motion for multiple robots, but they assume navigation objectives or motion goals rather than jointly predicting which robot should service each task and in which order. FlockDiffusion addresses this complementary problem by encoding the drone-task scene as a graph, predicting capacity-feasible ordered task bundles, and using those bundles as explicit conditions for parallel fleet trajectory generation. Few-step DDIM sampling and parallel fleet denoising then remove the search-based trajectory planner from the learned deployment pipeline.

% We build on
% this to replace the entire inference stack of prior system with a single learned
% generator (Fig.~\ref{fig:arch}): a graph neural network encodes drones, tasks, and
% obstacles and drives an explicit \emph{allocation head} predicting the task to drone
% assignment, an \emph{assignment conditioned} diffusion transformer samples all fleet
% trajectories in parallel, and a closed form decoder converts the samples into smooth,
% jerk limited routes with complete service and exact task coordinates guaranteed by
% construction. No graph search runs at inference; the classical Dijkstra and A star
% pipeline is used only offline to build the expert dataset. Factorizing which drone
% serves which tasks from how each drone flies makes the diffusion problem nearly
% unimodal, so it samples in very few steps. 

In summary, our contributions are:
\begin{itemize}
  \item A \textbf{diffusion-based fleet trajectory generator} that combines a
  scene graph neural network, an explicit allocation head, assignment
  conditioned parallel denoising, and a closed form decoder. The model performs
  no graph search at inference, runs 6 to 8 times faster than the classical
  stack, and generates shorter, smoother routes with complete task service.

  \item An \textbf{autoregressive task bundling teacher} built on the 
  Centralized Training with Decentralized Execution (CTDE) framework and trained using
  Proximal Policy Optimization (PPO). Our redesign introduces a lightweight local task
  multilayer perceptron, recurrent bundle memory, capacity-aware sequential
  selection, and vectorized bundle refinement. Bundling increases task
  completion from 50\% to 100\%, while the complete teacher further reduces
  route cost by 8.4\% relative to MAGNNET with bundling.

  \item \textbf{Real-time fleet demonstrations} in PyBullet and Gazebo across
  environments with varied obstacle geometry, showing feasible, collision-free
  execution of the generated multi-drone trajectories.
\end{itemize}

% ----------------------------------------------------------------------
\section{RELATED WORK}
\label{sec:related_work}

\subsection{Reinforcement Learning for Task Allocation}
\label{subsec:related_rl_allocation}

Multi-robot task allocation is inherently combinatorial because assignment quality depends on agent state, task geometry, capacity, and the future cost of servicing additional objectives. Reinforcement learning has consequently been used to replace manually designed allocation heuristics with policies that adapt their decisions to the current fleet and task configuration.
RTAW~\cite{agrawal2023rtaw} formulates warehouse task allocation as a Markov decision process and uses an attention-based PPO policy whose representation does not depend directly on the number of robots or tasks. This improves scalability, although allocation remains separated from the trajectories executed by the robots. DC-MRTA~\cite{agrawal2022dcmrta} couples an RL allocator with decentralized navigation through feedback from the navigation layer, allowing assignment decisions to account for executable motion. Nevertheless, task allocation and path generation are still performed by distinct modules.

Several approaches introduce additional structure to handle large allocation spaces. DL-DRL~\cite{xiao2024dldrl} separates multi-UAV scheduling into an upper-level task assignment policy and a lower-level route construction policy. This hierarchical factorization supports large problem instances but requires separate learned decision stages. Paul et al.~\cite{paul2024bigraph} retain a classical bipartite matching solver while using graph reinforcement learning to predict robot-to-task edge weights, demonstrating the value of combining learned representations with explicit optimization. MAGNNET ~\cite{ratnabala2025magnnet}, the closest basis for our allocation teacher, combines a graph neural network with a multi-agent PPO policy to represent agent-to-task interactions and produce decentralized assignments. Its deployment pipeline, however, still depends on separately computed motion costs and search-based path execution.

These methods demonstrate that reinforcement learning and graph representations can produce effective assignments, but they generally preserve a \emph{plan-then-fly} decomposition. Assignment is predicted first, after which another planner constructs the trajectories. In contrast, FlockDiffusion uses the predicted and refined assignment as an explicit conditioning signal for joint fleet trajectory generation. Task allocation and flight geometry are therefore coupled within the learned inference process.

\subsection{Diffusion Models for Planning and Allocation}
\label{subsec:related_diffusion}

Denoising diffusion models generate structured samples by iteratively transforming noise into data~\cite{ho2020ddpm}. DDIM reduces the required sampling steps~\cite{song2021ddim}, while Diffusion Transformers provide a scalable architecture for high-dimensional denoising ~\cite{peebles2023dit}. In planning, Diffuser generates complete trajectories conditioned on objectives~\cite{janner2022diffuser}, and Diffusion Policy models multimodal robot actions through conditional denoising ~\cite{chi2023diffusionpolicy}. These approaches primarily address single-agent planning or control.

Diffusion has also been applied to combinatorial optimization. DIFUSCO~\cite{sun2023difusco} performs graph-based denoising to generate discrete solutions for problems such as the traveling salesperson problem. However, discrete assignments or tours do not directly provide smooth, obstacle-aware multi-robot trajectories. FlockDiffusion connects these settings by predicting task-to-drone assignments from a scene graph and using them to condition the parallel generation of continuous fleet trajectories. This jointly captures task ownership, visit geometry, and trajectory completion within a unified learned framework.

\subsection{Reinforcement Learning Supervision for Diffusion}
\label{subsec:related_rl_diffusion}

Reinforcement learning and diffusion can be combined in several ways. One line of work optimizes a diffusion policy directly using reward feedback~\cite{black2024training}. Another uses reinforcement learning to generate or improve demonstrations and then distills those demonstrations into a diffusion model. The latter approach moves expensive exploration and optimization offline while retaining efficient diffusion inference at deployment.

DiffusionRL~\cite{makarova2025diffusionrl} follows this teacher-to-student pattern for robotic grasping. Reinforcement learning is used to adapt and improve a large grasp dataset, after which a lightweight diffusion policy is trained on the resulting task-aligned demonstrations. X-Sim ~\cite{dan2025xsim} similarly trains an RL policy in simulation using object-centric rewards and then distills its synthetic rollouts into an image-conditioned diffusion policy. These methods show that RL-generated data can transfer optimized behavior into a diffusion model without requiring the RL policy to remain in the deployed control loop.

% FlockDiffusion adopts this principle for multi-drone task bundling and trajectory generation. A graph-based PPO allocation policy and classical execution stack act as an offline teacher, grouping spatially compatible tasks into capacity-feasible bundles and producing their ordered expert trajectories. The student jointly learns an allocation head and an assignment-conditioned diffusion model from these demonstrations. At inference, the teacher is removed: the scene graph is encoded once, predicted assignments are refined into ordered bundles so that each drone services multiple tasks along its route, and the conditioned denoiser generates all fleet trajectories in parallel. Unlike prior RL-to-diffusion methods focused on manipulation, our formulation transfers route-aware multi-agent bundling together with continuous trajectory generation, while the decoder guarantees complete service, capacity compliance, and exact task visitation.

FlockDiffusion adopts this teacher-to-student principle for multi-drone task bundling and trajectory generation. A graph-based PPO allocation policy and classical execution stack act as an offline teacher, grouping spatially compatible tasks into capacity-feasible ordered bundles and producing their corresponding trajectory demonstrations. The student jointly learns an allocation head and an assignment-conditioned diffusion model from these demonstrations. At inference, the teacher is removed: the scene graph is encoded once, predicted assignments are refined into ordered bundles, and the conditioned denoiser generates all fleet trajectories in parallel. Unlike prior RL-to-diffusion approaches focused primarily on manipulation, our formulation transfers route-aware multi-agent allocation together with continuous fleet trajectory generation. The final decoder preserves the assigned task coordinates and visit sequence by construction, while capacity and unique task ownership are inherited from the refined bundles.

% --- Architecture figure placeholder ---
\begin{figure*}[t]
  \centering
  \includegraphics[width=\linewidth]{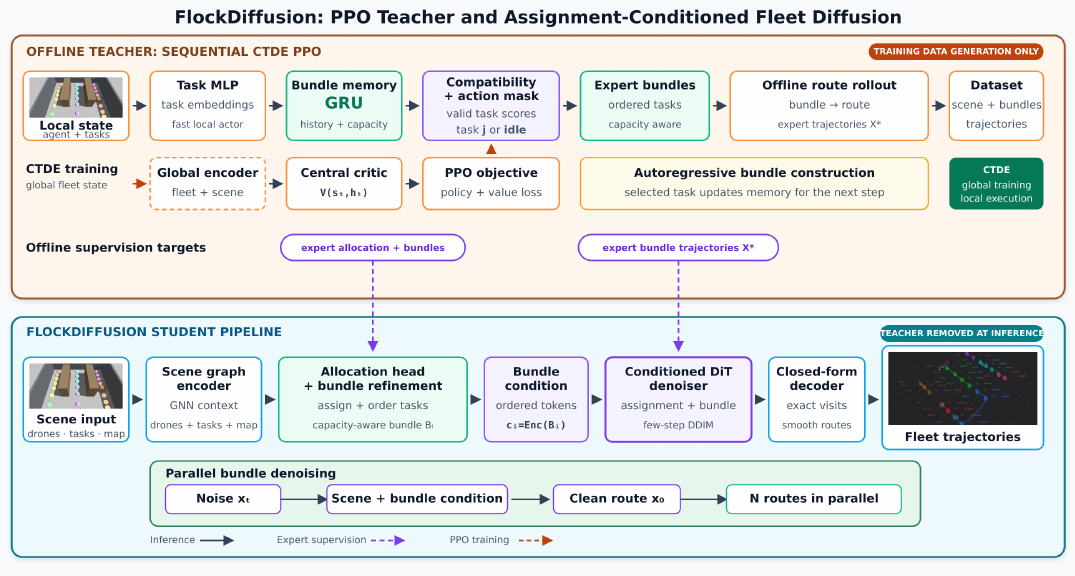}
  \caption{\textbf{FlockDiffusion training and inference architecture.}
The sequential CTDE PPO teacher uses a lightweight task encoder, recurrent bundle memory, and masked task selection to construct ordered task bundles. A centralized critic supplies the PPO training signal, while an offline route rollout converts the bundles into expert fleet trajectories. These expert allocations, bundles, and trajectories supervise the student. The student encodes the drone, task, and obstacle scene, predicts and refines capacity feasible task bundles, and conditions a diffusion transformer on the resulting ordered bundle tokens. The transformer denoises all drone trajectories in parallel, and a closed form decoder enforces exact task visits and produces smooth fleet routes. The teacher is used only during offline data generation and is removed at inference.}

  \label{fig:flockdiffusion_method}
\end{figure*}
 
% FlockDiffusion methodology and architecture figure.
% Suggested packages: amsmath, amssymb, graphicx.

% FlockDiffusion methodology and architecture figure.
% Suggested packages: amsmath, amssymb, graphicx.

% FlockDiffusion methodology and architecture figure.
% Suggested packages: amsmath, amssymb, graphicx.

\section{METHOD}
\label{sec:method}

\subsection{Problem Formulation}
\label{sec:problem_formulation}

We consider a fleet of $N$ drones operating in a bounded three-dimensional workspace containing $M$ tasks and a set of obstacles. Drone $i$ starts at $\mathbf{s}_i\in\mathbb{R}^{3}$ and can service at most $C_i$ tasks. Each task $j$ is represented by a location $\mathbf{g}_j\in\mathbb{R}^{3}$ and must be visited by exactly one drone. The allocation is represented by $\mathbf{A}\in\{0,1\}^{N\times M}$, where $A_{ij}=1$ indicates that drone $i$ services task $j$. For each drone, the assigned tasks are arranged into an ordered bundle
\begin{equation}
    \mathcal{B}_i=(b_{i,1},\ldots,b_{i,K_i}),
    \qquad K_i\leq C_i,
\end{equation}
where $b_{i,r}$ is the index of the $r$th task visited by drone $i$, $r\in\{1,\ldots,K_i\}$, and $K_i=\sum_{j=1}^{M}A_{ij}$ is its bundle size. The fleet trajectory is $\mathbf{X}=\{\mathbf{X}_i\}_{i=1}^{N}$, with $\mathbf{X}_i\in\mathbb{R}^{T\times 3}$, where $T$ is the number of sampled positions per drone and each row contains three spatial coordinates. Indices $i\in\{1,\ldots,N\}$ and $j\in\{1,\ldots,M\}$ refer to drones and tasks, respectively. A valid solution must satisfy \begin{equation}
    \sum_{i=1}^{N} A_{ij}=1 \quad \forall j,
    \qquad
    \sum_{j=1}^{M} A_{ij}\leq C_i \quad \forall i,
\end{equation}
The first constraint assigns each task exactly once, and the second bounds each drone's workload by $C_i$. A valid solution must also produce smooth, obstacle-aware trajectories that visit the tasks in the order specified by each bundle. Allocation and trajectory generation are therefore coupled: the quality of a bundle depends on the flight geometry required to execute it.

FlockDiffusion addresses this coupling through offline teacher supervision and a single learned student pipeline. A sequential policy trained using Proximal Policy Optimization (PPO) under Centralized Training with Decentralized Execution (CTDE) first generates expert task bundles and executable trajectories. The student then learns both the discrete allocation structure and the continuous fleet geometry from these demonstrations. Figure~\ref{fig:flockdiffusion_method} summarizes the complete training and inference procedure.

\subsection{Separation from MAGNNET.}
% The method contains an offline teacher adapted from MAGNNET~\cite{ratnabala2025magnnet} and a newly trained FlockDiffusion student. From MAGNNET, we retain the graph based global representation, CTDE formulation, PPO optimization objective, centralized critic, and discrete task or idle action space. We redesign the decentralized actor: instead of running graph processing for every local decision, our actor uses a lightweight multilayer perceptron (MLP) to encode the currently available tasks. We further add a recurrent bundle state and an autoregressive, capacity aware construction procedure in which every selected task updates the context for the next decision. An offline route rollout converts the resulting ordered bundles into trajectory demonstrations. The entire teacher branch is used only to generate supervision.

The offline teacher is adapted from MAGNNET~\cite{ratnabala2025magnnet}, whereas the FlockDiffusion student is trained independently from its generated demonstrations. From MAGNNET, the teacher retains the graph-based global representation, CTDE formulation, PPO objective, centralized critic, and discrete task-or-idle action space. We redesign the decentralized actor by replacing per-decision graph processing with a lightweight MLP that encodes the currently available tasks. We further introduce a recurrent bundle state and an autoregressive, capacity-aware construction procedure in which each selected task updates the context for the next decision. An offline route rollout then converts the resulting ordered bundles into trajectory demonstrations. The complete teacher branch is used only for offline data generation.

The lower branch in Fig.~\ref{fig:flockdiffusion_method} is the new FlockDiffusion model. Its scene graph encoder is distinct from both the teacher's lightweight local MLP and the MAGNNET graph module. The local MLP embeds task candidates independently for fast sequential teacher decisions, whereas the student scene graph neural network (GNN) performs relational message passing across drones and tasks and appends obstacle condition tokens. Thus, graph reasoning is retained where global scene structure is useful, while the local teacher actor uses our cheaper MLP design. The student scene GNN is trained jointly with a new allocation head, bundle refinement module, bundle conditioner, diffusion transformer, and closed form trajectory decoder. None of the teacher actor or critic parameters are used by the student at inference. 

\subsection{Autoregressive CTDE PPO Teacher}
\label{sec:ppo_teacher}

The upper branch of Fig.~\ref{fig:flockdiffusion_method} combines the graph based CTDE PPO foundation inherited from MAGNNET with our lightweight local MLP actor and autoregressive bundle construction. The local actor does not use the FlockDiffusion student scene GNN. At allocation step $k$, drone $i$ receives a local observation $\mathbf{o}_{i}^{k}$ containing its current state, the available task descriptors, the remaining capacity, and a valid action mask. A lightweight multilayer perceptron maps every candidate task feature $\mathbf{f}_{j}^{k}$ to an embedding
\begin{equation}
    \mathbf{e}_{j}^{k}=\phi_{\mathrm{task}}(\mathbf{f}_{j}^{k}).
\end{equation}
Here, $k$ indexes allocation decisions, $\mathbf{f}_{j}^{k}$ is the feature vector of task $j$, $\phi_{\mathrm{task}}$ is the learned task MLP, and $\mathbf{e}_{j}^{k}$ is its output embedding. The task embeddings are combined with an agent context embedding $\mathbf{z}_{i}^{k}$, obtained from the local observation $\mathbf{o}_{i}^{k}$, and a recurrent bundle state $\mathbf{h}_{i}^{k}$. The recurrent state is updated as
\begin{equation}
    \mathbf{h}_{i}^{k}
    =\operatorname{GRU}\!\left(
        \mathbf{h}_{i}^{k-1},
        [\mathbf{z}_{i}^{k},\mathbf{e}_{a_{i}^{k-1}}^{k}]
      \right),
\end{equation}
where $\operatorname{GRU}$ is a gated recurrent unit, $\mathbf{h}_{i}^{k-1}$ is its previous hidden state, and brackets denote feature concatenation. The index $a_{i}^{k-1}$ identifies the previously selected task, so $\mathbf{e}_{a_{i}^{k-1}}^{k}$ denotes its task embedding at the current decision step. This expression describes updates following a task selection; an initial or idle action requires a designated non-task input to the recurrent unit. This memory records the allocation history, occupied bundle slots, and remaining capacity. The actor computes an agent to task compatibility score
\begin{equation}
    \ell_{ij}^{k}
    =\psi\!\left(\mathbf{z}_{i}^{k},
                  \mathbf{h}_{i}^{k},
                  \mathbf{e}_{j}^{k}\right),
\end{equation}
where $\psi$ is the learned compatibility function and $\ell_{ij}^{k}$ is the scalar, unnormalized score for assigning task $j$ to drone $i$ at step $k$. The actor then applies the action mask, which excludes invalid choices, before selecting a valid task or the idle action. The selected task is appended to $\mathcal{B}_i$ and becomes part of the next recurrent context. This process continues until all tasks are allocated or all valid bundle slots are filled.

The teacher follows centralized training with decentralized execution. During training, a centralized critic receives the global fleet state, the task set, and the environment context, and estimates $V_{\omega}(\mathbf{S}^{k},\mathbf{h}^{k})$, where $V_{\omega}$ is the value function with learned parameters $\omega$, $\mathbf{S}^{k}$ denotes the global system state, and $\mathbf{h}^{k}=\{\mathbf{h}_{i}^{k}\}_{i=1}^{N}$ collects the fleet's recurrent states. Its output estimates expected discounted future reward. The global state $\mathbf{S}^{k}$ is distinct from the fixed start position $\mathbf{s}_i$. The actor is optimized with the clipped PPO objective~\cite{ratnabala2025hippo}, while execution uses only the local observation and recurrent bundle memory. Consequently, the teacher benefits from a global learning signal without requiring global processing in the local allocation policy.

After a complete set of bundles has been produced, an offline execution planner rolls out each ordered bundle in the training environment. Each demonstration contains the scene $\mathcal{G}^{*}$, expert assignments $\mathbf{A}^{*}$, ordered bundles $\mathcal{B}^{*}$, and resampled fleet trajectories $\mathbf{X}^{*}\in\mathbb{R}^{N\times T\times 3}$. A superscript $*$ marks teacher demonstrations, and $\mathcal{B}^{*}=\{\mathcal{B}_i^{*}\}_{i=1}^{N}$ collects their ordered bundles. The scene representation $\mathcal{G}^{*}$ contains the associated drone, task, and obstacle information. The teacher and the offline execution planner are used only to construct this dataset and are removed from the deployment pipeline.

\subsection{Scene Graph Encoding and Bundle Prediction}
\label{sec:scene_bundle_prediction}

This subsection describes the new FlockDiffusion student rather than the MAGNNET-derived PPO teacher. The student receives the drone states, task states, and obstacle description. We represent the scene as a heterogeneous graph whose valid drone and task nodes exchange information through message passing. For layer $l$, a node representation is updated according to
\begin{equation}
    \mathbf{h}_{u}^{l+1}
    =\sigma\!\left(
      \mathbf{W}_{\mathrm{s}}^{l}\mathbf{h}_{u}^{l}
      +\frac{1}{|\mathcal{N}(u)|}
       \sum_{v\in\mathcal{N}(u)}
       \mathbf{W}_{\mathrm{n}}^{l}\mathbf{h}_{v}^{l}
    \right).
\end{equation}
Here, $u$ and $v$ index graph nodes, $l$ indexes message passing layers, and $\mathbf{h}_{u}^{l}$ is the feature embedding of node $u$ at layer $l$. This graph embedding is distinct from the teacher's recurrent state $\mathbf{h}_{i}^{k}$. The set $\mathcal{N}(u)$ contains the neighbors of $u$, and $|\mathcal{N}(u)|$ is its cardinality. $\mathbf{W}_{\mathrm{s}}^{l}$ and $\mathbf{W}_{\mathrm{n}}^{l}$ are learned linear maps for self and neighbor features, respectively, and $\sigma$ is the elementwise activation. The neighbor term is taken as zero for an isolated node. Drone to task edges communicate assignment compatibility, while drone to drone edges provide fleet wide context about position and load. Obstacle features are encoded as additional condition tokens and concatenated with the graph tokens. The resulting scene representation is denoted by $\mathbf{H}_{\mathcal{G}}$, the collection of condition tokens for scene $\mathcal{G}$. In our implementation, three message passing rounds operate at hidden width 128, after which the valid node features are projected to 256 dimensional condition tokens. Obstacle features are processed by a separate lightweight encoder and appended to the graph representation. With ten drone slots, twenty task slots, and eight obstacle slots, the encoder produces at most 38 scene condition tokens.

An allocation head predicts a task to drone distribution as:
\begin{equation}
    p_{ij}=p(A_{ij}=1\mid\mathbf{H}_{\mathcal{G}}).
\end{equation}

Here, $p_{ij}$ is the predicted probability that drone $i$ serves task $j$, conditional on the scene tokens; $\sum_{i=1}^{N}p_{ij}=1$ for each valid task. The predicted assignment is refined into ordered, capacity-feasible bundles. This refinement removes duplicate service, enforces each drone capacity, and orders the tasks assigned to every drone. The resulting bundle $\widehat{\mathcal{B}}_i$ is encoded into bundle tokens:
\begin{equation}
    \mathbf{c}_{i}^{\mathrm{B}}
    =\phi_{\mathrm{B}}(\widehat{\mathcal{B}}_i,
                       \mathbf{H}_{\mathcal{G}}).
\end{equation}

Here, a hat denotes a predicted or refined quantity, $\phi_{\mathrm{B}}$ is the bundle encoding function, and $\mathbf{c}_{i}^{\mathrm{B}}$ contains the condition tokens for drone $i$'s ordered bundle; the superscript $\mathrm{B}$ denotes bundle conditioning. Conditioning on the ordered bundle explicitly separates the combinatorial decision of which tasks a drone serves from the continuous decision of how it flies. This reduces ambiguity in the trajectory distribution and supports accurate sampling with few denoising steps. 
The allocation head is supervised using the expert teacher assignments:
\begin{equation}
    \mathcal{L}_{\mathrm{alloc}}
    =-\frac{1}{M}\sum_{j=1}^{M}\sum_{i=1}^{N}
       A_{ij}^{*}\log p_{ij}.
\end{equation}

This cross entropy loss $\mathcal{L}_{\mathrm{alloc}}$ averages over the $M$ tasks. The binary teacher label $A_{ij}^{*}$ selects the expert drone for task $j$, and $\log$ denotes the natural logarithm. 

\subsection{Assignment Conditioned Fleet Diffusion}
\label{sec:assignment_conditioned_diffusion}

The diffusion model learns the distribution of executable fleet trajectories conditioned on the scene and ordered task bundles. During training, Gaussian noise is applied to the normalized expert trajectory $\mathbf{X}^{*}$ using a forward diffusion process~\cite{ho2020ddpm}
\begin{equation}
    q(\mathbf{X}_{t}\mid\mathbf{X}^{*})
    =\mathcal{N}\!\left(
       \sqrt{\bar{\alpha}_{t}}\mathbf{X}^{*},
       (1-\bar{\alpha}_{t})\mathbf{I}
     \right).
\end{equation}
Here, $t\in\{1,\ldots,S\}$ indexes diffusion noise levels, where $S$ is the number of training diffusion steps; $t$ is distinct from a flight sample index and $S$ is distinct from the trajectory length $T$. $\mathbf{X}_{t}$ is the noisy fleet trajectory, and $q$ denotes the forward noising distribution. The symbol $\mathcal{N}(\boldsymbol{\mu},\boldsymbol{\Sigma})$ denotes a Gaussian distribution with mean $\boldsymbol{\mu}$ and covariance $\boldsymbol{\Sigma}$, distinct from the graph neighborhood notation. The cumulative signal coefficient is $\bar{\alpha}_{t}=\prod_{r=1}^{t}(1-\beta_r)$, where $\beta_r$ is the noise variance at diffusion step $r$. The identity matrix $\mathbf{I}$ acts on the flattened trajectory coordinates, giving independent noise in each coordinate. We use clean trajectory prediction. The denoiser therefore estimates
\begin{equation}
    \widehat{\mathbf{X}}_{0}
    =f_{\theta}\!\left(
       \mathbf{X}_{t},t,
       \mathbf{H}_{\mathcal{G}},
       \mathbf{c}^{\mathrm{B}}
     \right),
\end{equation}
where $f_{\theta}$ is the denoising network with learned parameters $\theta$, $\widehat{\mathbf{X}}_{0}$ is its estimate of the clean fleet trajectory, and $\mathbf{c}^{\mathrm{B}}=\{\mathbf{c}_{i}^{\mathrm{B}}\}_{i=1}^{N}$ collects the bundle conditions for the entire fleet. The subscript $0$ denotes the clean diffusion state, not the initial flight position. The trajectory is divided into temporal patches and combined with learned drone and time embeddings. A diffusion transformer processes the scene tokens and all drone trajectory tokens jointly, enabling every generated path to account for the other agents and the shared environment. All $N$ trajectories are denoised in parallel rather than planned independently. We use $T=64$ trajectory samples and patches of four time steps, yielding 16 trajectory tokens per drone. The denoiser contains eight transformer blocks with width 256, eight attention heads, and adaptive layer normalization with zero initialized residual gates. 

The diffusion loss is the active drone masked mean squared error
\begin{equation}
    \mathcal{L}_{\mathrm{diff}}
    =\frac{1}{T\sum_{i=1}^{N} m_i}
      \sum_{i=1}^{N}m_i
      \left\|
        \widehat{\mathbf{X}}_{0,i}-\mathbf{X}_{i}^{*}
      \right\|_{\mathrm{F}}^{2},
\end{equation}
where $m_i\in\{0,1\}$ equals one for an active drone and zero for an inactive or padded slot. The slice $\widehat{\mathbf{X}}_{0,i}$ is the predicted clean trajectory of drone $i$, and $\mathbf{X}_i^{*}$ is its normalized expert trajectory. The squared Frobenius norm $\|\cdot\|_{\mathrm{F}}^{2}$ sums squared errors across all $T$ samples and three coordinates. Thus, $\mathcal{L}_{\mathrm{diff}}$ averages squared position error over active drone samples, assuming at least one active drone. The complete student loss is
\begin{equation}
    \mathcal{L}
    =\lambda_{\mathrm{alloc}}\mathcal{L}_{\mathrm{alloc}}
     +\lambda_{\mathrm{diff}}\mathcal{L}_{\mathrm{diff}}.
\end{equation}
Here, $\mathcal{L}$ is the combined student training objective, and $\lambda_{\mathrm{alloc}}$ and $\lambda_{\mathrm{diff}}$ are nonnegative weights balancing allocation and trajectory supervision. At inference, the reverse process is evaluated with deterministic Denoising Diffusion Implicit Model (DDIM) sampling~\cite{song2021ddim}. The direct clean trajectory parameterization, assignment conditioning, deterministic updates, and parallel fleet prediction allow the model to use a small number of sampling steps. Training uses $S=1000$ steps with a cosine diffusion schedule and an exponential moving average of the denoiser parameters; only the short DDIM trajectory is evaluated at deployment.

\subsection{Constraint Preserving Trajectory Decoding}
\label{sec:trajectory_decoding}

The sampled trajectories provide smooth global flight geometry, while the decoder enforces the discrete service structure. For drone $i$, its start and the ordered task coordinates in $\widehat{\mathcal{B}}_i$ are treated as fixed anchors. The generated samples provide intermediate control points between these anchors. A centripetal Catmull--Rom construction converts the control points into a continuous spatial curve, and a quintic minimum jerk time law
\begin{equation}
    \rho(\tau)=10\tau^{3}-15\tau^{4}+6\tau^{5},
    \qquad \tau\in[0,1],
\end{equation}
produces smooth progress along each segment. Here, $\tau$ is normalized time within a segment, and $\rho(\tau)$ is its normalized progress parameter. For segment duration $\Delta>0$ and elapsed time $u\in[0,\Delta]$, $\tau=u/\Delta$. The function moves from $\rho(0)=0$ to $\rho(1)=1$ with zero first and second derivatives at both endpoints. Since every assigned task is inserted as an exact anchor, the decoded routes preserve task coordinates and visit order by construction. Capacity and unique service are inherited from the refined bundles. Obstacle inflation and local geometric correction are applied during decoding, followed by collision validation in the execution environment.

The deployed pipeline therefore consists only of scene encoding, bundle prediction, parallel DDIM sampling, and closed form decoding. No teacher policy or graph search is evaluated online. 

\section{EXPERIMENTS AND RESULTS} 
\subsection{Evaluation metrics.}

We report \emph{planner coverage} (percentage of tasks included in generated plans) and \emph{execution completion} (percentage actually serviced during simulation). \emph{Planned route cost} captures aggregate fleet routing cost, while \emph{tasks per unit distance} measures service efficiency. For bundling methods, we additionally report \emph{bundle compactness} (spatial spread of co-assigned tasks) and \emph{mean marginal insertion cost} (added route cost per inserted task), with lower values preferred for both. \emph{Allocation episode steps} and the \emph{standard deviation of tasks per drone} characterize decision effort and workload balance, respectively. Computational cost is measured via allocation or diffusion generation time, A* and oracle time, and their summed total (including repair where applicable, excluding model loading); note that diffusion denoising iterations are distinct from allocation episode steps.
% \subsection{Ablation Studies} \label{sec:ablations}
\subsection{Teacher Ablation Study}\label{sec:ablations}
We conduct the ablation study in PyBullet, using controlled task and obstacle configurations to compare the task allocation policies. The baseline is the MAGNNET allocator~\cite{ratnabala2025magnnet}, while the second variant adds only the proposed task bundling mechanism. Our complete teacher additionally uses a lightweight local task MLP, recurrent bundle memory, autoregressive bundle construction, and vectorized refinement.

\begin{table}[t]
    \centering
    \caption{Teacher architecture ablation in PyBullet. Arrows indicate the
    preferred direction and -- denotes a metric that is not applicable.}
    \label{tab:teacher_ablation}
    \small
    \setlength{\tabcolsep}{3pt}
    \renewcommand{\arraystretch}{1.08}
    \begin{tabular}{@{}lrrr@{}}
        \toprule
        Metric & Baseline & \shortstack{Baseline +\\ bundling} & Ours \\
        \midrule
        Completion (\%) $\uparrow$
            & 50 & \textbf{100} & \textbf{100} \\
        Route cost $\downarrow$
            & 413.10 & 338.80 & \textbf{310.22} \\
        Bundle compactness $\downarrow$
            & -- & 0.47 & \textbf{0.45} \\
        Mean insertion cost $\downarrow$
            & -- & 12.90 & \textbf{10.84} \\
        Tasks / distance $\uparrow$
            & -- & 0.06 & \textbf{0.07} \\
        Episode steps $\downarrow$
            & 100.00 & 4.20 & \textbf{3.78} \\
        Tasks / robot std.
            & 0.00 & \textbf{0.97} & 1.01 \\
        \bottomrule
    \end{tabular}
\end{table}

Table~\ref{tab:teacher_ablation} shows that adding bundling to the baseline increases task completion from 50\% to 100\%. Our complete teacher preserves 100\% completion while reducing route cost by 8.4\%, mean marginal insertion cost by 16.0\%, and episode steps by 10.0\% relative to the baseline with bundling. Bundle compactness improves from 0.47 to 0.45, and tasks per unit distance increase from 0.06 to 0.07. These results show that the complete teacher architecture provides additional routing benefits beyond the bundling mechanism alone.

Bundle compactness, marginal insertion cost, and tasks per distance describe the quality of constructed bundles. They are therefore not applicable to the original baseline, which does not perform task bundling. The zero workload dispersion reported for the baseline must also be interpreted together with its 50\% completion rate. Our complete teacher has slightly higher workload dispersion than the baseline with bundling, with a standard deviation of 1.01 rather than 0.97 tasks per robot.

The quantitative ablation results in Table~\ref{tab:teacher_ablation} are
obtained exclusively in PyBullet.

\subsection{Experimental Setup}
% The experiments were carried out in a customized Crazyflie simulation environment built on ROS2 Humble, Crazyswarm2, Crazyflie firmware software-in-the-loop (SITL), and Gazebo Harmonic. Each simulated Crazyflie ran its own firmware instance and communicated with the Crazyswarm2 ROS interface through a dedicated UDP port. This setup provided pose feedback and allowed position commands to be sent through the same multi-robot software stack used throughout the experiments. The simulator also supported adding and removing tasks during operation, introducing drone failures, automatically replanning assignments, and executing the resulting trajectories. These features allowed the task-allocation methods to be tested and evaluated under changing conditions.
 % This simulation assess the feasibility of deploying the complete allocation and trajectory generation pipeline in a physics based simulation environment. Moreover, it allows to evaluate scalability of the framework.

The experiments were conducted in a customized Crazyflie simulation environment built on ROS~2 Humble, Crazyswarm2, Crazyflie firmware software-in-the-loop (SITL), and Gazebo Harmonic. Each simulated Crazyflie ran an independent firmware instance and communicated with the Crazyswarm2 ROS interface through a dedicated UDP port. The setup provided pose feedback and accepted position commands through the same multi-robot software stack used for the fleet experiments. The simulator additionally supports dynamic task insertion and removal, simulated drone failures, assignment replanning, and trajectory execution. In this work, it is used to evaluate the complete allocation and trajectory-generation pipeline in physics-based simulation and to study computational scaling with task density.

\subsection{Qualitative Results}
\begin{figure*}[h]
    \centering
    \includegraphics[width=\textwidth]{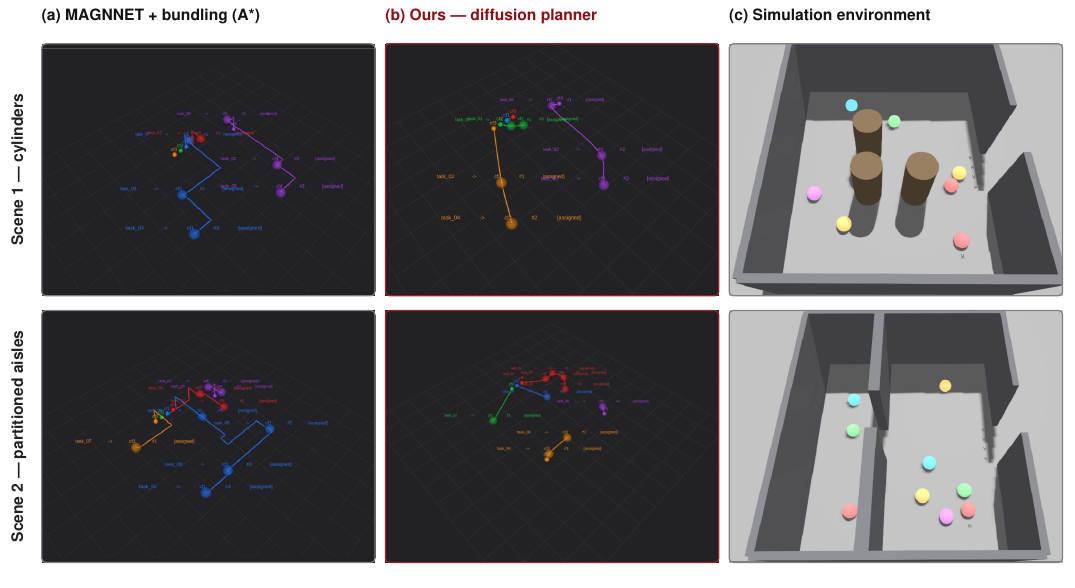}
    \caption{\textbf{Qualitative multi-drone allocation and trajectory results.}
  Column (a) shows MAGNNET~+ bundling with A* execution and column (b) our
  diffusion planner, both visualizing the ordered task bundles and generated
  fleet trajectories; column (c) shows the corresponding Gazebo scene. Rows
  are two environments: cylindrical obstacles (top) and a constrained
  partitioned-aisle corridor (bottom). Colors associate each drone with the
  tasks assigned along its route, illustrating multi task collection under
  different obstacle geometries.}
    \label{fig:qualitative_results}
\end{figure*}
% We use Gazebo for the qualitative
% demonstrations in Fig.~\ref{fig:qualitative_results}, where the generated task
% bundles and fleet trajectories are executed in environments with different
% obstacle geometries.
% Preamble: \usepackage{graphicx}
% Upload the images/ directory and insert \input{qualitative-results}.

% Suggested qualitative-results paragraph:
Figure~\ref{fig:qualitative_results} illustrates the generated allocations and fleet trajectories in two environments with distinct geometric constraints. In the cylindrical scene, the assigned task sequences must traverse the free space between multiple obstacles, while in the corridor scene the routes organize tasks around narrow partitioned passages. In both examples, each colored route connects several tasks assigned to the same drone, qualitatively demonstrating the task bundling behavior learned by FlockDiffusion. The comparison against MAGNNET~+ bundling also makes the effect of the diffusion planner apparent: whereas A* execution turns only on voxel axes and produces axis-aligned staircase segments with sharp heading reversals at every waypoint, jointly denoising all trajectories yields curvature-continuous paths that pass through the same task sequence without stopping to turn. Smoother paths are directly preferable for aerial platforms: eliminating the stop-and-turn behavior at each waypoint removes the repeated deceleration and reacceleration that dominate energy use on a quadrotor, keeps commanded accelerations within the vehicle's actuation limits, and leaves trajectories that the onboard tracking controller can follow with far lower position error.
% \subsection{Simulation Results}

% Preamble: \usepackage{booktabs}

\subsection{Gazebo Pipeline Evaluation}
\label{sec:gazebo_evaluation}

We evaluate the complete allocation and trajectory generation pipelines across five Gazebo environments, each containing ten drones and 20 tasks. Both MAGNNET variants use A* for trajectory planning, while FlockDiffusion uses a diffusion model and decoder. These experiments extend the teacher policy ablations conducted in PyBullet.

\begin{table}[t]
    \centering
    \caption{Average pipeline results across five Gazebo scenes.
    Completion denotes planner coverage. Allocation time includes
    trajectory generation for diffusion. Planning time includes the
    logged A* and oracle stages; total is the sum of reported allocation,
    planning, and repair times, excluding model loading.
    -- denotes not applicable.}
    \label{tab:gazebo_pipeline}
    \small
    \setlength{\tabcolsep}{3pt}
    \renewcommand{\arraystretch}{1.06}
    \begin{tabular}{@{}lrrr@{}}
        \toprule
        Metric & MAGNNET &
        \shortstack{MAGNNET\\+ bundling} &
        \shortstack{Flock\\Diffusion} \\
        \midrule
        Trajectory planner & A* & A* & Diffusion \\
        Plan completion (\%) $\uparrow$
            & 50 & \textbf{100} & \textbf{100} \\
        Planned route cost $\downarrow$
            & 394.40 & 192.80 & \textbf{163.20} \\
        Bundle compactness $\downarrow$
            & -- & 0.362 & \textbf{0.279} \\
        Tasks / distance $\uparrow$
            & 0.026 & 0.105 & \textbf{0.123} \\
        Episode steps $\downarrow$
            & 1000 & 4 & \textbf{1} \\
        \midrule
        Alloc./generation (ms) $\downarrow$
            & 4473.58 & \textbf{8.40} & 24.28 \\
        A* + oracle (ms) $\downarrow$
            & \textbf{30.12} & 36.50 & 53.86 \\
        Total stage time (ms) $\downarrow$
            & 4503.70 & \textbf{44.90} & 78.14 \\
        \bottomrule
    \end{tabular}
\end{table}

Table~\ref{tab:gazebo_pipeline} shows that FlockDiffusion achieves 100\% planner coverage in every scene, matching MAGNNET with bundling and exceeding the original baseline's 50\%. Relative to MAGNNET with bundling, it reduces mean planned route cost by 15.4\%, improves bundle compactness by 22.9\%, and increases tasks per unit distance by 17.3\%. It produces the fleet plan in one recorded allocation step, compared with four for the bundling baseline and 1000 for MAGNNET. This single allocation step contains multiple diffusion denoising iterations.

Diffusion allocation and trajectory generation average 24.28\,ms, compared with 4473.58\,ms for MAGNNET and 8.40\,ms for MAGNNET with bundling; the large latency of the original baseline accompanies its 1000-step rollouts and incomplete allocation. 

\paragraph{Latency analysis.}
The timing breakdown shows that FlockDiffusion's higher latency arises primarily from trajectory generation rather than task assignment. Its allocation head requires only 1.29\,ms, while embedding and five denoising steps require 17.65\,ms and trajectory decoding adds 5.33\,ms. In comparison, MAGNNET with bundling spends 8.40\,ms on policy rollout and 2.08\,ms on A* planning. The Gazebo evaluation also records higher oracle time for diffusion, increasing from 34.42 to 53.86\,ms. Consequently, the reported stage sum is 78.14\,ms for FlockDiffusion versus 44.90\,ms for MAGNNET with bundling. 
% These measurements indicate
% a tradeoff in the evaluated configurations: diffusion reduces planned
% route cost by 15.4\%, but its iterative generation and decoding exceed
% the cost of the baseline's inexpensive A* stage. The additional oracle
% overhead requires profiling before attributing it to the diffusion
% architecture.

% Preamble: \usepackage{graphicx}

\begin{figure}[t]
    \centering
    \includegraphics[width=0.8\columnwidth]{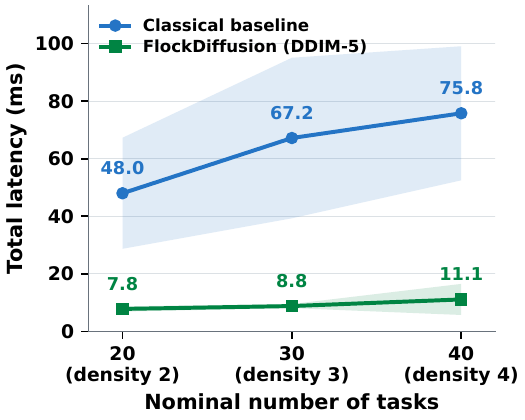}
    \caption{\textbf{Latency scaling with task density.}
    Mean total latency over 100 scenes per density for ten drones.
    Nominal task counts are 20, 30, and 40, with approximately
    17, 26, and 34 realized tasks, respectively. Shaded bands
    show the reported variation around each mean.}
    \label{fig:latency_scalability}
\end{figure}

\paragraph{Scalability.}
Figure~\ref{fig:latency_scalability} evaluates the optimized pipeline as nominal task counts increase from 20 to 40. FlockDiffusion's mean latency rises from 7.8 to 11.1\,ms, compared with 48.0 to 75.8\,ms for the classical baseline. Across the three densities, FlockDiffusion achieves a 6.2 to 7.6 times speedup, retaining a 6.8 times advantage at the highest density. Fixed token dimensions, five denoising steps, and vectorized bundle refinement limit computational growth, demonstrating favorable latency scaling over the tested range. This benchmark isolates the same allocation head, denoising, and decoding stages profiled in Section~\ref{sec:gazebo_evaluation}, where trajectory generation, not the allocation head, accounts for most of FlockDiffusion's latency.
The scalability benchmark and Gazebo experiments use separate evaluation protocols. The former measures the optimized pipeline across task densities, whereas the latter reports timings from the deployed simulation implementation, including separately logged oracle processing. Their absolute latencies are therefore reported separately and are not directly comparable. 

\section{Conclusion}
We presented FlockDiffusion, a framework combining an autoregressive bundling teacher with assignment conditioned diffusion for multi-drone task allocation and trajectory generation. PyBullet ablations showed that bundling increased task completion from 50\% to 100\%, while our complete teacher further reduced route cost by 8.4\% relative to MAGNNET with bundling. The optimized scalability benchmark achieved 6.2 to 7.6 times faster inference than the classical pipeline, with latency increasing from 7.8 to 11.1\,ms as nominal task counts increased from 20 to 40 for ten drones. In the separate Gazebo planning evaluation, FlockDiffusion achieved 100\% task coverage and reduced planned route cost by 15.4\%. These simulations demonstrate efficient coordination of 10 drones under increasing task density, enabling evaluation of configurations that are logistically demanding to reproduce with physical fleets.

\bibliographystyle{IEEEtran}
\bibliography{references}

\end{document}